\documentclass[Afour,sageh,times]{sagej}

\newcommand{\MONTH}{%
  \ifcase\the\month
  \or January% 1
  \or February% 2
  \or March% 3
  \or April% 4
  \or May% 5
  \or June% 6
  \or July% 7
  \or August% 8
  \or September% 9
  \or October% 10
  \or November% 11
  \or December% 12
  \fi}
  
\usepackage{xspace}
\usepackage{times}
\usepackage{epsfig}
\usepackage{graphicx}
\usepackage{amsmath}
\usepackage{amssymb}
\usepackage[english,algoruled,linesnumbered,vlined]{algorithm2e}
\let\oldnl\nl% Store \nl in \oldnl
\newcommand{\nonl}{\renewcommand{\nl}{\let\nl\oldnl}}% Remove line number for one line

\usepackage[font=footnotesize]{caption}

\usepackage[caption=false]{subfig} % for subfigures
\usepackage{wrapfig}
\usepackage{float}
\usepackage{tabu,multirow}
\restylefloat{table}

\usepackage[table,xcdraw]{xkeyval,xcolor}

\usepackage{siunitx} % used for degrees symbol

\newlength{\colwidth}
\newcommand{\todo}[1]{}
\renewcommand{\todo}[1]{{\color{red} TODO: {#1}}}

\usepackage{manyfoot}

\makeatletter
\DeclareRobustCommand\onedot{\futurelet\@let@token\@onedot}
\def\@onedot{\ifx\@let@token.\else.\null\fi\xspace}

\def\eg{\emph{e.g}\onedot} 
\def\ie{\emph{i.e}\onedot} 
\def\cf{\emph{c.f}\onedot} 
\def\etc{\emph{etc}\onedot} \def\vs{\emph{vs}\onedot}
\def\wrt{w.r.t\onedot} 
\def\etal{\emph{et al}\onedot}

\makeatother

\makeatletter
\newlength{\sfp@hseplen}\newlength{\sfp@vseplen}
\define@cmdkey{subfigpos}[sfp@]{pos}[ul]{}% \sfp@pos
\define@cmdkey{subfigpos}[sfp@]{font}[\small]{}% \sfp@font
\define@cmdkey{subfigpos}[sfp@]{vsep}[2\baselineskip]{\setlength{\sfp@vseplen}{\sfp@vsep}}% \sfp@vsep
\define@cmdkey{subfigpos}[sfp@]{hsep}[10pt]{\setlength{\sfp@hseplen}{\sfp@hsep}}% \sfp@hsep
\newcommand{\subfigimg}[3][,]{%
  \setkeys{Gin,subfigpos}{pos,font,vsep,hsep,#1}% Set (default) keys
  \setbox1=\hbox{\includegraphics{#3}}% Store image in box
  \ifnum\pdfstrcmp{\sfp@pos}{ul}=0% UPPER LEFT placement of subfig label
    \leavevmode\rlap{\usebox1}% Print image
    \rlap{\hspace*{\sfp@hsep}\raisebox{\dimexpr\ht1-\sfp@vsep}{\sfp@font{#2}}}% Print label
    \phantom{\usebox1}% Insert appropriate spacing
  \else\ifnum\pdfstrcmp{\sfp@pos}{ur}=0% UPPER RIGHT placement of subfig label
    \leavevmode\usebox1% Print image
    \llap{\raisebox{\dimexpr\ht1-\sfp@vsep}{\sfp@font{#2}}\hspace*{\sfp@hsep}}% Print label
  \else\ifnum\pdfstrcmp{\sfp@pos}{lr}=0% LOWER RIGHT placement of subfig label
    \leavevmode\usebox1% Print image
    \llap{\raisebox{\sfp@vsep}{\sfp@font{#2}}\hspace*{\sfp@hsep}}% Print label
  \else% Assume LOWER LEFT placement of subfig label
    \leavevmode\rlap{\usebox1}% Print image
    \rlap{\hspace*{\sfp@hseplen}\raisebox{\sfp@vsep}{\sfp@font{#2}}}% Print label
    \phantom{\usebox1}% Insert appropriate spacing
  \fi\fi\fi
}
\makeatother

\usepackage{stackengine}

\usepackage{moreverb,url}

\usepackage[colorlinks,bookmarksopen,bookmarksnumbered,citecolor=red,urlcolor=red]{hyperref}

\newcommand\BibTeX{{\rmfamily B\kern-.05em \textsc{i\kern-.025em b}\kern-.08em
T\kern-.1667em\lower.7ex\hbox{E}\kern-.125emX}}

\def\volumeyear{2019}

\begin{document}

\runninghead{Tsesmelis, Hasan, Cristani, Del Bue and Galasso}

\title{An integrated light management system with real-time light measurement and human perception}

\author{Theodore Tsesmelis\affilnum{1,2,3}, Irtiza Hasan\affilnum{1,2,3}, Marco Cristani\affilnum{3}, Alessio Del Bue\affilnum{2,}$^{\dagger}$ and Fabio Galasso\affilnum{1,}$^{\dagger}$}

\affiliation{\affilnum{1}OSRAM Licht AG, DE\\
\affilnum{2}Istituto Italiano di Tecnologia (IIT), IT\\
\affilnum{3}University of Verona (UNIVR), IT\\
\affilnum{$^{\dagger}$}Equal contribution}

\corrauth{Theodore Tsesmelis, OSRAM Licht AG
Innovation \& Innoventures,
Vision Lab,
Parkring 33,
85748 Garching, DE.}

% \email{alistair.smith@sunrise-setting.co.uk}

\begin{abstract}

Illumination is important for well-being, productivity and safety across several environments, including offices, retail shops and industrial warehouses. Current techniques for setting up lighting require extensive and expert support and need to be repeated if the scene changes. Here we propose the first fully-automated light management system (LMS) which measures lighting in real-time, leveraging an RGBD sensor and a radiosity-based light propagation model. Thanks to the integration of light distribution and perception curves into the radiosity, we outperform a commercial software (Relux) on a newly introduced dataset. Furthermore, our proposed LMS is the first to estimate both the presence and the attention of the people in the environment, as well as their light perception. Our new LMS adapts therefore lighting to the scene and human activity and it is capable of saving up to 66\%, as we experimentally quantify, without compromising the lighting quality.
\end{abstract}

\keywords{Machine vision, light modeling, radiosity, illumination map, light management system, human-centric lighting}

\maketitle

\section{Introduction}
% % The very first letter is a 2 line initial drop letter followed
% % by the rest of the first word in caps.
% % 
% % form to use if the first word consists of a single letter:
% % \IEEEPARstart{A}{demo} file is ....
% % 
% % form to use if you need the single drop letter followed by
% % normal text (unknown if ever used by the IEEE):
% % \IEEEPARstart{A}{}demo file is ....
% % 
% % Some journals put the first two words in caps:
% % \IEEEPARstart{T}{his demo} file is ....
% % 
% % Here we have the typical use of a "T" for an initial drop letter
% % and "HIS" in caps to complete the first word.
% for IEEE journal papers produced under \LaTeX\ using
% IEEEtran.cls version 1.8b and later.
% % You must have at least 2 lines in the paragraph with the drop letter
% % (should never be an issue)
% I wish you the best of success.

 

% \subsection{Subsection Heading Here}
% Subsection text here.

% % needed in second column of first page if using \IEEEpubid
% %\IEEEpubidadjcol

% \subsubsection{Subsubsection Heading Here}
% Subsubsection text here.

Light is indispensable in our perception of the world and it affects our emotional and physiological responses~\cite{partonen2000bright, kuller1993melatonin}.
%Good-quality lighting can improve well-being, visual performance and interpersonal communications.
Well-lit workplaces provide visual comfort and improve productivity \cite{kralikova2016lighting,boyce2004lighting}. However, lighting may reach 15\% of the overall building electricity consumption, with peaks above 25\% \cite{kralikova2015energy}. To save energy, a light management system (LMS) would need to measure lighting and to reason on the human perception of it in real-time, \eg to dim down lighting where none sees it.

Current LMSs cannot automatically obtain dense measures of the spatial illumination in real-time.
The former task is either accomplished manually with luxmeters (point-to-point measurements) or via offline CAD-based simulations. These require an operator to visit the scene and perform the measurements, or a detailed CAD model of the 3D scene, respectively. Both approaches need to be repeated upon any scene change and are impractical in the flexible open-plan spaces of modern offices, designed to adapt to the daily work plan.

Modern LMSs cannot either estimate the human perception of light. Current implementations rely on daylight harvesting and occupancy sensing. The first adjusts the luminaires to maximize the use of daylight, when available \cite{Kaminska2018}. The second leverages thermal- or radar-based motion detectors~\cite{guo2010performance} to switch on/off all luminaires in the room when people enter/leave, no matter the size of the (open-plan) office \cite{de2017occupancy}.

Here we introduce a new LMS which estimates both the scene lighting and the human perception of it, to optimize lighting and its quality while saving energy. Light estimation leverages an RGBD camera and a radiosity model for light propagation, and distinguishes the scene 3D structure, the object reflectance and light positions. For human light perception, we locate people in the environment, estimate their visual frustum of attention (VFOA) and the incident light onto their VFOA, depending on their position and gaze. We also introduce a new labelled dataset, featuring a number of lit rooms, with and without human activity. We provide RGBD images and 3D meshes, labelled with material reflectance properties; luminaire positions, characteristics and dimming level; person locations and VFOA. Light intensity and human perception are ground-truthed by luxmeters, placed in the scene or worn by people respectively.

Finally, we introduce a new end-to-end system architecture and implement the autonomous system that we call the \emph{``invisible light switch''} (ILS). \textit{ILS} encompasses an RGBD camera, a processor, a light controller, a communication bus and the luminaires. Since \textit{ILS} estimates how much light each person receives, it may switch off or dim those luminaires which are not visible, \eg on the other side of large open spaces or behind cubicle panels.
This removes the need for manual switches and provides a boost in energy efficiency, saving up to 66\% without compromising the light quality.

Our main contributions are: \textbf{\textit{i}}.\ we propose a real-time light estimation from an RGBD sensor as well as its perception by the scene occupants; \textbf{\textit{ii}}.\ we collect a new benchmark for quantitative evaluation; \textbf{\textit{iii}}.\ we propose an end-to-end autonomous LMS, to control lighting and save energy.
This manuscript brings together two previous conference publications on the topics of light estimation and control \cite{tsesmelis2018b,tsesmelis2018a}, and extends the work with \textbf{\textit{a}})\ novel dataset and annotations, \textbf{\textit{b}})\ more experiments, and \textbf{\textit{c}})\ the definition and implementation of the overall end-to-end system architecture.

We present the system and modelling in Section~\ref{sec:method}; We discuss the dataset and experiments in Section~\ref{sec:experiments}; Finally, Section~\ref{sec:conclusion} concludes the manuscript. Next we review related work.

% \begin{figure}[thbp]
% % \vspace{-10pt}
% 	\begin{center}
% 		\subfloat[][Light quality framework \cite{veitch2000lighting}.]{\includegraphics[width=0.49\linewidth]{pics/lighting_quality.png}\label{fig:light_quality}}
% 		\hspace{0.01em}
% 		\subfloat[][Energy saving strategies \cite{tak2009}.]{\includegraphics[width=0.49\linewidth]{pics/strategies.PNG}\label{fig:strategies}}\\
% % 		\hspace{0.01em}
% % 		\subfloat[][]
% % 		{\includegraphics[width=0.32\linewidth]{images/unit_sphere_over_cad.png}\label{fig:isocel_rays}}
%         \vspace{5pt}
%         % \decoRule
% 		\caption[Light Quality - Light Strategies]{\textbf{(a)} The framework of lighting quality according to the International Association of Lighting Designers (IALD) \cite{iald2018}. \textbf{(b)} From the bottom to the top this implies the exploitation of natural light, the distribution and control of lighting locally, efficient space configuration with bright colors and open spaces for better light propagation and utilization of the light sources driven by energy saving customization. The order of the strategies in the pyramid shows the importance that each action should be applied in a green-oriented building maintenance.}
% 		\label{fig:light_quality_and_energy_management}
% 	\end{center}
% % 	\vspace{-10pt}
% \end{figure}

\begin{figure*}[!ht]
	\begin{center}
	    \includegraphics[width=1\linewidth]{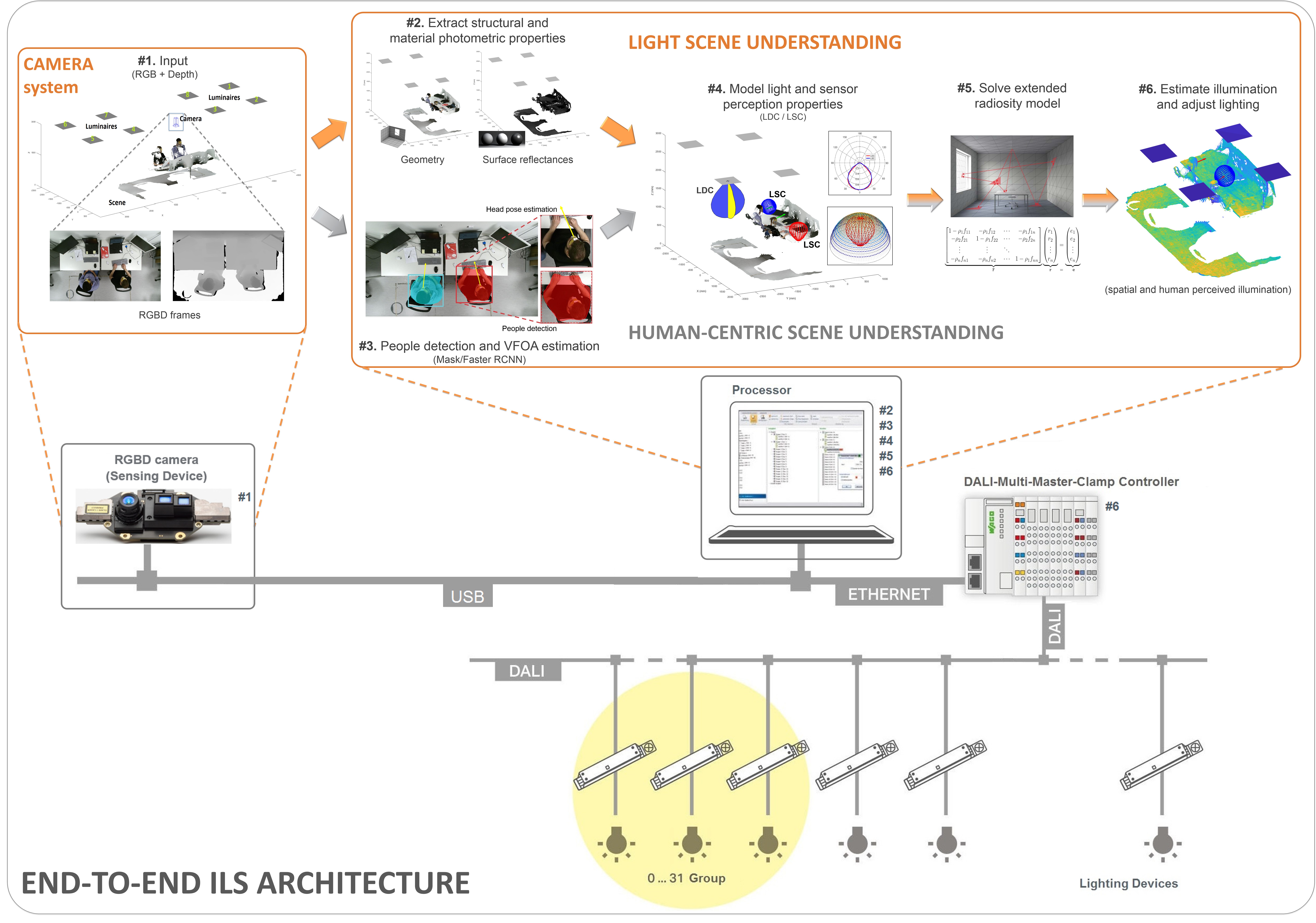}\\
	   % \vspace{5pt}
	   % \includegraphics[width=0.8\linewidth]{pics/dali_configuration_a(1).png}
	   % \decoRule
	\end{center}
	\vspace{-5pt}
	\caption[RGBD2Lux Pipeline]{
	We propose the invisible light switch (\textit{ILS}). This is the first end-to-end light management system (LMS) which understands the scene lighting and the people visual perception of it, towards a lighting efficiency application. The architecture includes an RGB-D sensor, installed in the room ceiling, which acquires top-view color and depth images, as detailed in the camera system box (step \#1 in the pipeline). The input RGB-D images are processed in the processor. There it extracts the photometric properties of the surfaces and the geometric structure of the scene by leveraging the RGB and depth image pairs (\textit{step \#2}) while in parallel it detects the people and estimates the VFOA of each of them (the yellow arrows) (\textit{Step \#3}). Thereafter, it models the actual light sources (luminaires) and light-perceiving devices (luxmeters or people) by using the corresponding distribution curves LDC/LSC (\textit{step \#4}). Finally, it solves for the extended radiosity model (\textit{Step \#5}) and estimates the spatial and human-perceived illumination over the visible 3D scene (\textit{Step \#6}). Based on the light estimation, we steer lighting in the room via the WAGO DALI Configurator software. This encodes luminaire commands to a DALI Master Light Controller via TCP/IP, which it relays to the luminaires via the DALI BUS.}\label{fig:teaser1}
 	\vspace{-10pt}
\end{figure*}

\section{Related Work}

Light measurement and management encompass different fields in science and engineering, which we review here.

\subsection{Light Measurement}

According to Cuttle \etal \cite{Cuttle2010} the current status of lighting profession and lighting evaluation is on luminance based light assessing procedures, \ie estimating the light arriving at the eyes of a virtual observer. However, in most previous work light measurement and modelling refer mainly to research in image and visual computing. As we review in \cite{tsesmelis2018a}, previous research regarded image spatio-temporal and pixel-like approaches \cite{cai2016luminance,hiscocks2014measuring}, and the creation of photorealistic renderings \cite{marschner2015fundamentals}, not the actual spatial lighting measurement. By contrast, commercial light modelling software, \eg Relux, DIALux, AGi32, focuses on measuring light. They mainly target offline measurements and the evaluation of lighting solutions in a simulated environment. We build our work on \cite{tsesmelis2018a}, adopting computer vision techniques to measure lighting in real-time by means of a camera.

\subsection{Light Management Systems (LMSs)}
LMSs play a crucial role in reducing energy consumption in offices~\cite{ul2014review} and prior work mainly addressed daylight harvesting~\cite{Kaminska2018} and occupancy sensing systems (OSS)~\cite{guo2010performance,Pandharipande2012}.
In particular, \cite{ul2014review} emphasized the potential of OSSs for energy saving, surveying available technology and the performance compromise; \cite{guo2010performance} additionally focused on the implementation of occupancy sensors; \cite{Pandharipande2012} specifically considered an OSS based on ultrasonic and proposed improvements in user localization based on time-difference-of-arrival and transmission over multiple time slots.
However, none of the above addressed energy saving, nor distinguished among private, single-user and open-plan offices as one would intuitively argue for. 

Among related work considering open-plan offices, \cite{kim2013workspace} determined three layout categories:
% In both cases, 
% the energy saving in open-plan offices was studied with limited attention: in fact, they do not distinguish among private, single user offices and open-plan offices, despite a different treatment is intuitively needed. 
(a) cubicle layout with high partitions ($\sim$1.5m or higher), (b) cubicle layout with low partitions (up to $\sim$1.5m) and (c) open layout with no or limited partitions.
\cite{Wang2012} and \cite{Pandharipande2013} proposed and demonstrated by simulation an optimization algorithm to compromise energy efficiency and user-comfort, by considering user presence and daylight conditions. They did not address any occupancy sensing algorithm, nor a real-time light propagation model, which we propose here for a real end-to-end light management system.
Finally, the lack of a proper review study on LMSs in open-plan offices has been recently stressed by \cite{de2017occupancy}.
They envisioned modern LMSs, capable of dimming light according to how much of it is perceived by each person in the scene, by means of their visual frustum of attention. We realize this vision here with our ``\textit{invisible light switch}''.

\section{Proposed End-to-end System \label{sec:method}}
% \subsection{System Overview} \label{sec:scene_composition_analysis}

The proposed end-to-end system encompasses three functional tasks: the \emph{light-centric scene understanding}, the \emph{human-centric scene understanding}, and the \emph{light management}. These are depicted in Fig. \ref{fig:teaser1}.%, \ref{fig:teaser2} and \ref{fig:system_architecture} respectively.

Light-centric scene understanding is responsible to estimate the scene 3D geometry and the light propagation within the scene. For the light propagation, we adopt and extend the radiosity model to additionally model the actual light source (office luminaires) as well as the receiving sensor (luxmeters or viewing people).

Human-centric scene understanding stands for detecting the people in the scene and estimating their view frustum of attention (VFOA). We use the VFOA to quantify how much light a single person perceives from each light source.

Both the light-centric and human-centric scene understanding modules are necessary when it comes to open-plan offices, where multiple light sources create complex light patterns which act on moving employees. Finally, light management stands for automatically adjusting lighting in response to the people position and attention. This includes dimming or switching the luminaires off when people do not see them, as in our \textit{ILS} system. The first two are sensing tasks, performed in first place by the sole use of an RGBD camera. The third task is about light control and leverages an established light-communication BUS (DALI), to read statuses and send commands.
%The initial light source properties (\ie total amount of light sources, their location, luminous intensity, and type) are given. Else no user input is needed.

\subsection{Light-centric Scene Understanding} \label{sec:scene_composition_analysis}

\textit{ILS} targets the estimation of the scene 6-DoF illumination, \ie anywhere in the 3D space and from any 3D direction, by the sole use of RGBD images. Towards this goal, we define a procedure to retrieve the scene 3D surface from the point cloud, as well as the surface reflectance. We then use and extend radiosity to model the light interaction among 3D scene parts by introducing the distribution characteristics of the actual lights (luminaires) and sensing elements (the luxmeters or the people in the scene) \cf illustration in Fig.~\ref{fig:teaser1}.
%real-time, automatic procedure

\noindent\textbf{From RGBD images to surface and reflectance.}
Light propagation requires a scene described by patches, \ie 3D facets characterized by a surface and an orientation. But an RGBD camera only provides a sparse and noisy point cloud.

We set to recover the scene surfaces by first denoising (weighted median and bilateral filtering) the point cloud by means of the color and depth image pairs%~\cite{wmf2014,Paris2009}
. Then we reconstruct the surface as a 3D mesh of multiple scattered patches by using the Open3D library%~\cite{Zhou2018}
. Given the surface normals, \textit{ILS} estimates the albedo $\rho$ of each surface element by the first-order spherical-harmonics method same to the one adapted in \cite{tsesmelis2018a}. This requires multiple images of each surface patch under different illumination conditions (\eg alternating the lights from each of the scene luminaires). We attain the image-set by time-lapse recordings and by selecting/synthesizing single-light-source-lit images with the method in \cite{Lit2017}.
%The main goal is to reconstruct the surface from the point cloud, represent it as a pair of vertices and faces (also known as patches), associate each face with a scalar albedo value, describe the geometric relationship of each patch to all others and finally estimate an illumination value for each patch.

\noindent\textbf{The Radiosity Model.}
Given the scene 3D surface and the corresponding reflectance, \textit{ILS} estimates the illuminance of each surface patch with the radiosity model%\cite{cohen1993rri}
. Radiosity is adopted by most commercial light simulation software (Relux, DIALux, AGi32) because it describes the physical light propagation phenomenon and provides light estimations which are close to those measured by luxmeters, in \textit{lux}. %By contrast, computer graphics techniques target realistic effects and only the relative plausibility of patch illumination estimation, \eg with respect to neighboring ones.

Radiosity is a \emph{simple} linear model but it has two main limitations: it only models point light sources and it disregards the sensor sensitivity characteristics (\cf Fig.~\ref{fig:teaser1}(Step 3): light is emitted/absorbed differently depending on the incident angle).
The second aspect affects  all  current  commercial  light  planning software.
We address both aspects by introducing into an extended radiosity model the Light Distribution Curve (LDC) (to address the \emph{non-ideal} \emph{non-point}-light-source luminaires) and the Luxmeter Sensitivity Curve (LSC) (to model the light observer/sensor). As detailed in \cite{tsesmelis2018a}, the extended radiosity encodes the LDC and LSC non-linearities into the scene form factors, thus preserving the model linearity and minimally affecting computation. 
\subsection{Human-centric Scene Understanding} \label{sec:human_centric_analysis}

The main human features of a light management system are the localization of people, the estimation of their visual attention and the consequent estimation of their lighting perception in the scene. %We describe here algorithms to address the three aspects, given an RGBD ceiling-mounted camera.

We cast the localization of people as a human detection task from top-view imagery based on the state-of-the-art Mask R-CNN model with the ResNet-101 backbone.
%We attain best performance by pre-training the detector on the large MS-COCO dataset-\cite{lin2014microsoft} (80k training + 35k validation images) and fine-tuning on the recent top-view people dataset of \cite{DemirkusetalVISAPP17TopViewDet} (4459 training images, mainly featuring a single person). 

We denote the visual attention of people by their visual frustum of attention (VFOA), which we estimate with the model of \cite{hasan2017tiny}. This consists of a Faster R-CNN architecture with VGG16 backbone, extended by a VFOA branch for gaze estimation, taking as input the detected person bounding-box. As noted in \cite{hasan2017tiny}, the whole-person bounding box provides an important contextual cue, to complement the tiny heads from the top-view imagery.
%\mc{This VFOA model supported the prediction of trajectories and activities of people~\cite{Hasan2018MXLST}, which may further benefit this work, \eg in terms of system latency and well being}.
%paving the way to future extension of the present work, i.e., predicting human pose and position for a even better light management

The VFOA is cast as classifying the person viewing angle into quantized direction bins. We experimented with 4-quantized viewing directions (\emph{North}, \emph{West}, \emph{South}, \emph{East}), as well as with 8 (yielding a granularity of $45^{\circ})$. Moreover, attempting to estimate the VFOA by regression, under-performed the classification approach, based on the fact that it is easier to estimate a class label than the exact angle. 
%We train the neural network on new annotations of the top-view people dataset of \cite{DemirkusetalVISAPP17TopViewDet}.
%randomly partitioned the data into training and testing set, in a ratio of 70\%-30\% respectively.
Once we extract the detected 2D people positions and viewing angles, we map them onto the 3D space by means of depth to 3D mapping.

Finally, we model the light perceived by the people as the illumination reaching the person's field-of-view, as described by the 6-DoF head position and VFOA orientation in the scene. Each field-of-view is assumed to be a conic reception field. The arriving light follows the radiosity model of Sec. \nameref{sec:scene_composition_analysis}, the ray-casting simulation as described for the light sensitivity curve (LSC) in \cite{tsesmelis2018b}, and the consequent integration across the human field-of-view. Overall, each person's light perception is approximated as a light sensor alike the luxmeter, positioned between the eyes.

%Luxmeters are also used to acquire ground truth
%The rays are projected in the space as a uniform generated sequence over the unit sphere and weighted accordingly, based on the modelled luxmeter's LSC and in a similar pattern used for the estimation of the spatial illumination.
%The contribution of each patch to the total amount of lighting perceived by the occupant, is computed by estimating the percentage of rays intersecting that patch.

% \begin{figure}[!ht]
% % \vspace{-10pt}
% 	\begin{center}

% 		\subfloat[][]{\includegraphics[width=0.45\linewidth]{pics/lsc_a.png}\label{fig:lsc_a}}
% 		\hspace{0.01em}
% 		\subfloat[][]{\includegraphics[width=0.45\linewidth]{pics/lsc_modeling.png}\label{fig:lsc_modeling}}
% % 		\hspace{0.01em}
% % 		\subfloat[][]
% % 		{\includegraphics[width=0.32\linewidth]{images/unit_sphere_over_cad.png}\label{fig:isocel_rays}}
%         \vspace{5pt}
% 		\caption[Modeling of LSC from Human Perspective]{Modeling of the Luxmeter Sensitivity Curve (LSC) as a human light perception model.}
% 		\label{fig:lsc_sampling}
% 	\end{center}
% % 	\vspace{-10pt}
% \end{figure}

% The LSC illustrates the  perception characteristic of every luxmeter sensor which in this work we adopt in order to meet the measuring requirements of the collected ground truth data and to simulate the human light perception. We have chosen this solution because this is the standard de facto in the lighting industry and it provides satisfactory solutions when doing light commissioning \cite{ies2011commissioning}.

\subsection{Light Management System}\label{sec:lightmanagement}

% \begin{figure}[!ht]
% 	\begin{center}
% 	    \includegraphics[width=1\linewidth]{pics/dali_configuration_a.png}
% 	   % \subfloat[][]{\includegraphics[width=0.7\linewidth]{pics/system_architecture.png}}
% 	   % \hspace{0.01em}
% 	   % \subfloat[][]{\includegraphics[width=0.3\linewidth]{pics/dali_configuration.jpg}}
% 	   % \decoRule
% 	\end{center}
% 	\vspace{-5pt}
% 	\caption[]{
% 	The end-to-end \textit{ILS} architecture, as described in Sec.~\ref{sec:lightmanagement}. Note the RGBD sensor connected to the processor (exemplified as a laptop) via USB. The processor runs the algorithms to estimate the scene illumination and how much of it each person perceives. It also runs the WAGO DALI Configurator program to encode luminaire commands to a DALI Master Light Controller via TCP/IP. The controller relays the commands to the luminaire and reads out their statuses via the DALI BUS.
%     }\label{fig:system_architecture}
%  	\vspace{-5pt}
% \end{figure}

\textit{ILS} is a camera-aided smart LMS to control lighting in response to the currently spatial and human perceived estimated light as well as the people position and attention. The \textit{ILS} reads the luminaire status and switches them on and off based on the people presence. Additionally it dims luminaires down when partially visible by the people, therefore saving energy \emph{``in the invisible''}, while maintaining the desirable scene illumination.

As illustrated in Fig. \ref{fig:teaser1}, the proposed LMS consists of the sensing RGBD camera, a computing device to estimate the light and human factors in the scene (\cf Secs. \nameref{sec:scene_composition_analysis}), \nameref{sec:human_centric_analysis}, a Master Light Controller to communicate commands to luminaires, and the luminaires themselves, interconnected via a suitable protocol BUS. \textit{ILS} is implemented as a computer program running on the computing device, based on the people presence and attention, and on the luminaire status readouts. Overall, the proposed LMS system is autonomous, end-to-end and real-time.

We have adopted and report results on images acquired via a Kinect v2 RGBD camera. 
%, compared to competing sensors such as the Bluetechnix Argos.
The computing device is a laptop, running the sensing algorithms of Secs.~\nameref{sec:scene_composition_analysis}, \nameref{sec:human_centric_analysis} and the controlling invisible light switch program. The Kinect v2 is connected to a laptop via a USB2 port.

The Master Light Controller interfaces the computer with the luminaires, by forwarding switching and dimming commands and reporting the luminaire statuses (on, off, dimming level, \etc). We adopt the WAGO-I/O-SYSTEM 750 through a DALI (\emph{Digital Addressable Lighting Interface}) Multi-Master Module 753-647 \cite{dali753-647}. As with most such controllers, it connects to the computer via Ethernet TCP/IP, enabling therefore IoT and cloud-based intelligence. The connection is accommodated via proprietary software, \ie the WAGO DALI Configurator \cite{dali_configurator} running on the computer. The WAGO DALI Configurator allows easy commissioning of the devices connected on the DALI network. This includes the offline configuration of the entire DALI network, including the electromagnetic control gears (ECGs), the sensors and the saving/repeating device configurations.

As communication BUS between the Master Light Controller and the luminaires \textit{ILS} adopts the established DALI BUS. DALI is one of the simplest duplex-communication protocols, which allows to flexibly connect up to 64 devices in series, grouping them into up to 16 clusters.

\section{Experiments} \label{sec:experiments}

Here we evaluate the performance of the LMS system and its parts on a novel benchmark.
%First, we benchmark the accuracy of the scene illumination map (Sec.~\ref{sec:luxcom}) and of the detector and head-pose estimator (Sec.~\ref{sec:head_estimation_eval}). Then we evaluate the effectiveness of ILS (Sec.~\ref{sec:light_perception}). Next we illustrate the dataset and metrics.

% Experiments are organized as follows. Sec. \ref{sec:dataset} presents a novel dataset for light measurements. Sec. \ref{sec:luxcom} shows the results for the case of  spatial light estimation based on the different applied studies and describes in more details the evaluation of our radiosity model against state of the art approaches. Sec. \ref{sec:head_estimation_eval} reports the results regarding the person occupancy and head pose estimation study, while Sec. \ref{sec:light_perception} describes in more details the evaluation for the gaze-gathered light estimation. Finally, Sec. \ref{sec:applications} evaluates both the spatial and gaze-gathered light estimation as a power saving application.

\begin{figure}[!ht]
% \vspace{-5pt}
	\begin{center}
	    \includegraphics[width=1\linewidth]{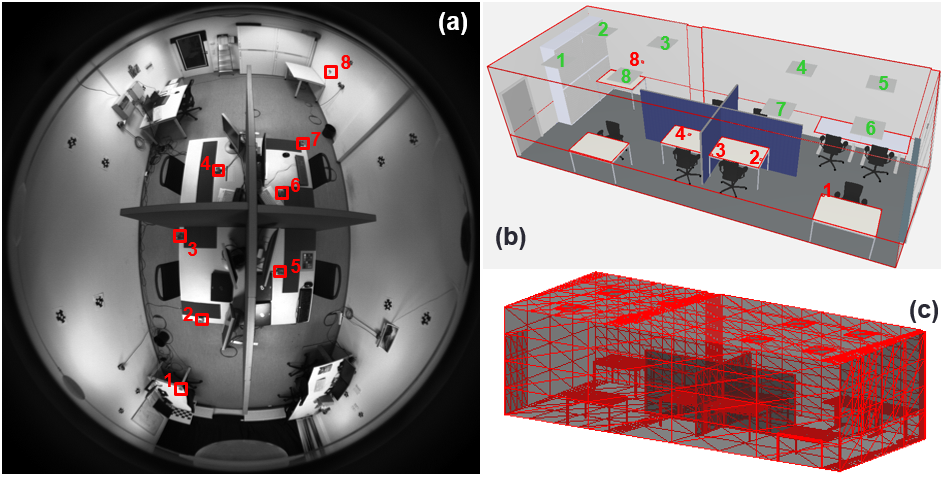}\\ %\hspace{.3cm}
% 		\includegraphics[width=0.48\linewidth]{images/render_room1_2.PNG} \\
% 		\textit{Room1}% \hspace{3cm} \textit{Room2}

        % \captionsetup{justification=centering}
        \vspace{-7pt}
        \captionsetup[subfigure]{position=bottom, labelformat=empty, textfont=normalfont, singlelinecheck=off, justification=raggedleft}
        \addtocounter{subfigure}{3}
		\subfloat[][]{\topinset{{\footnotesize(d)}}{\includegraphics[width=0.32\linewidth]{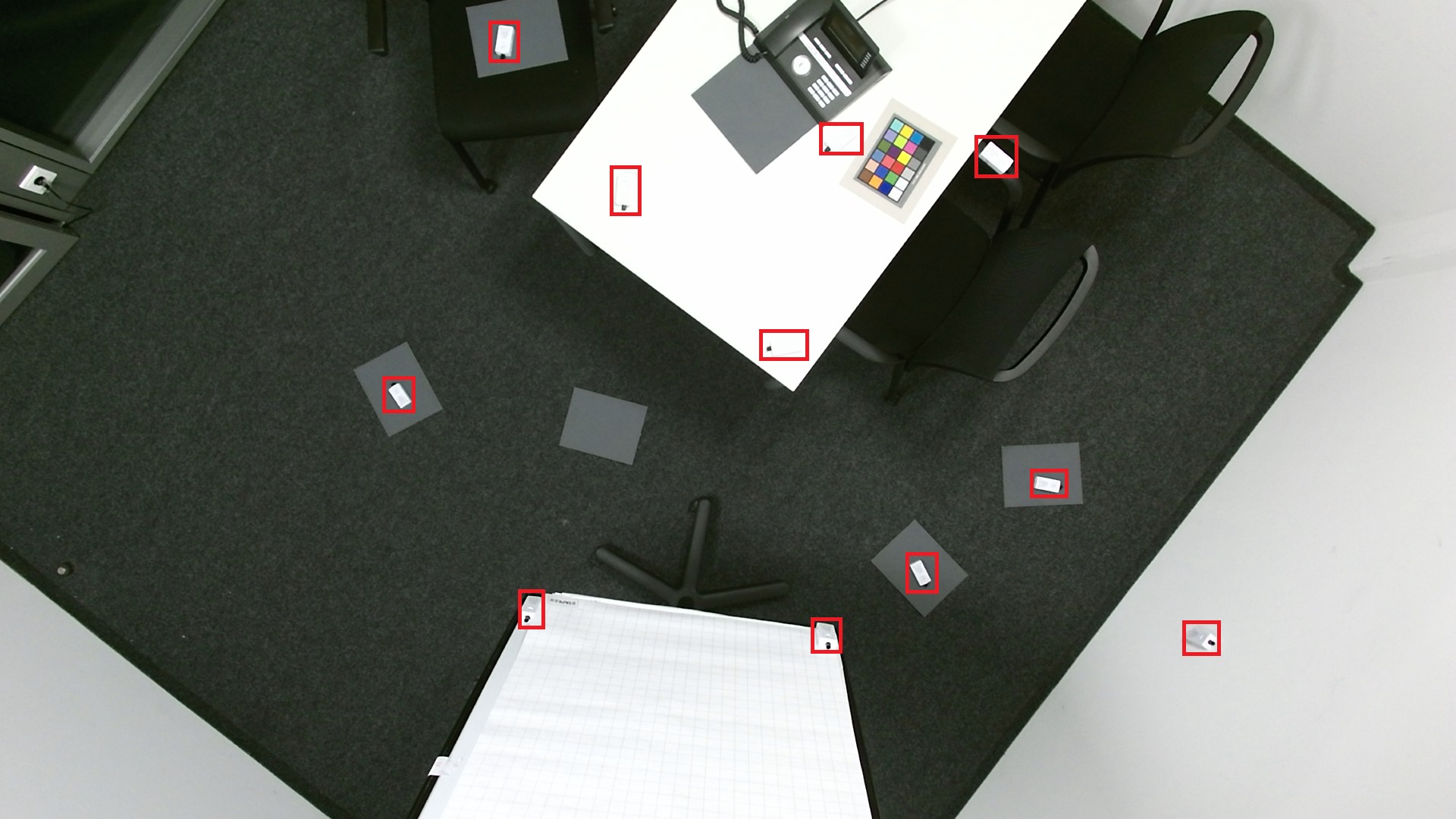}}{37pt}{70pt}\label{fig:room1}}
		\hspace{0.01em}
% 		\captionsetup{justification=centering}
		\subfloat[][]{\topinset{{\footnotesize(e)}}{\includegraphics[width=0.32\linewidth]{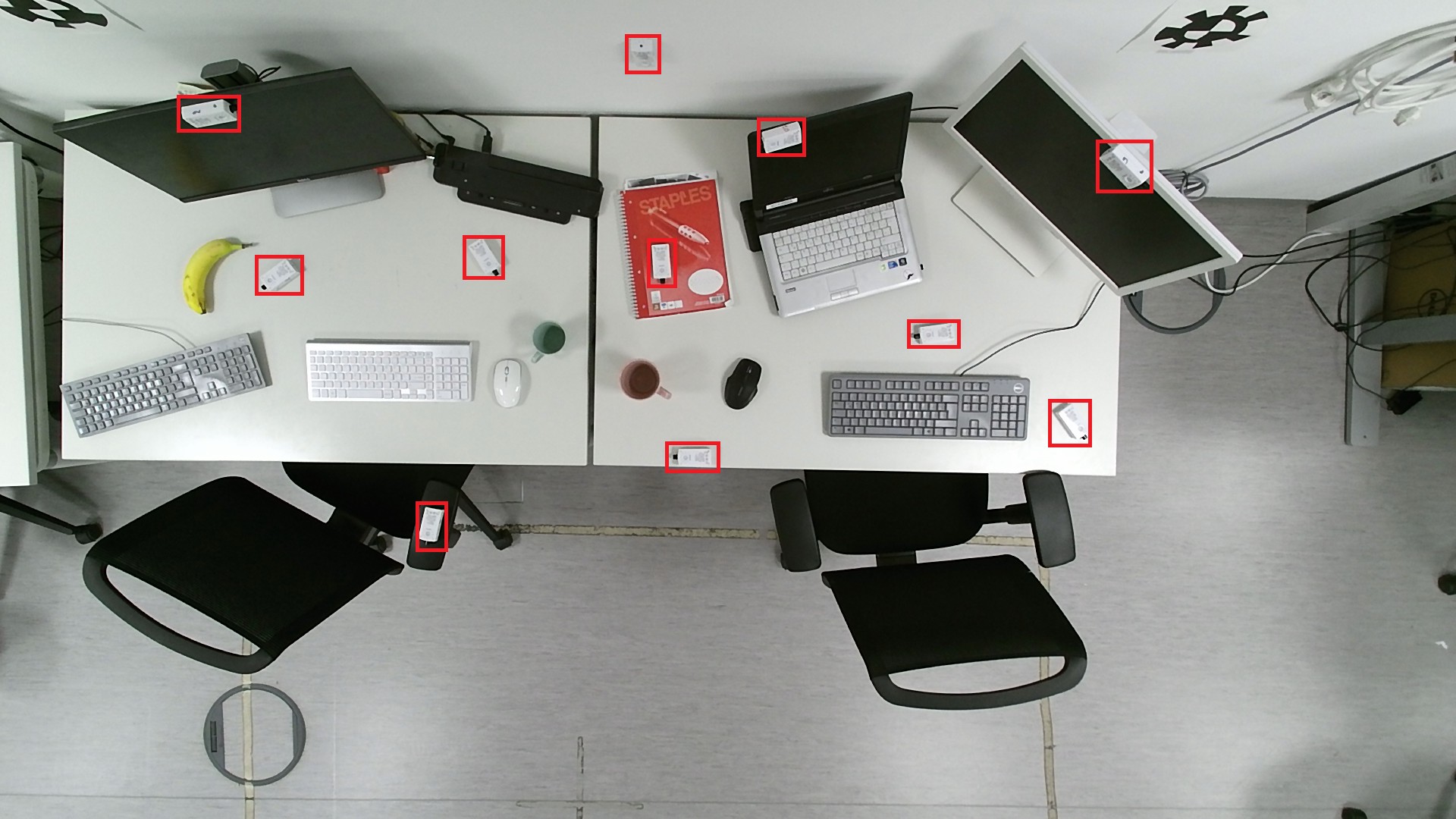}}{37pt}{70pt}\label{fig:room4}}
		\hspace{0.01em}
% 		\captionsetup{justification=centering}
		\subfloat[][]{\topinset{{\footnotesize(f)}}{\includegraphics[width=0.32\linewidth]{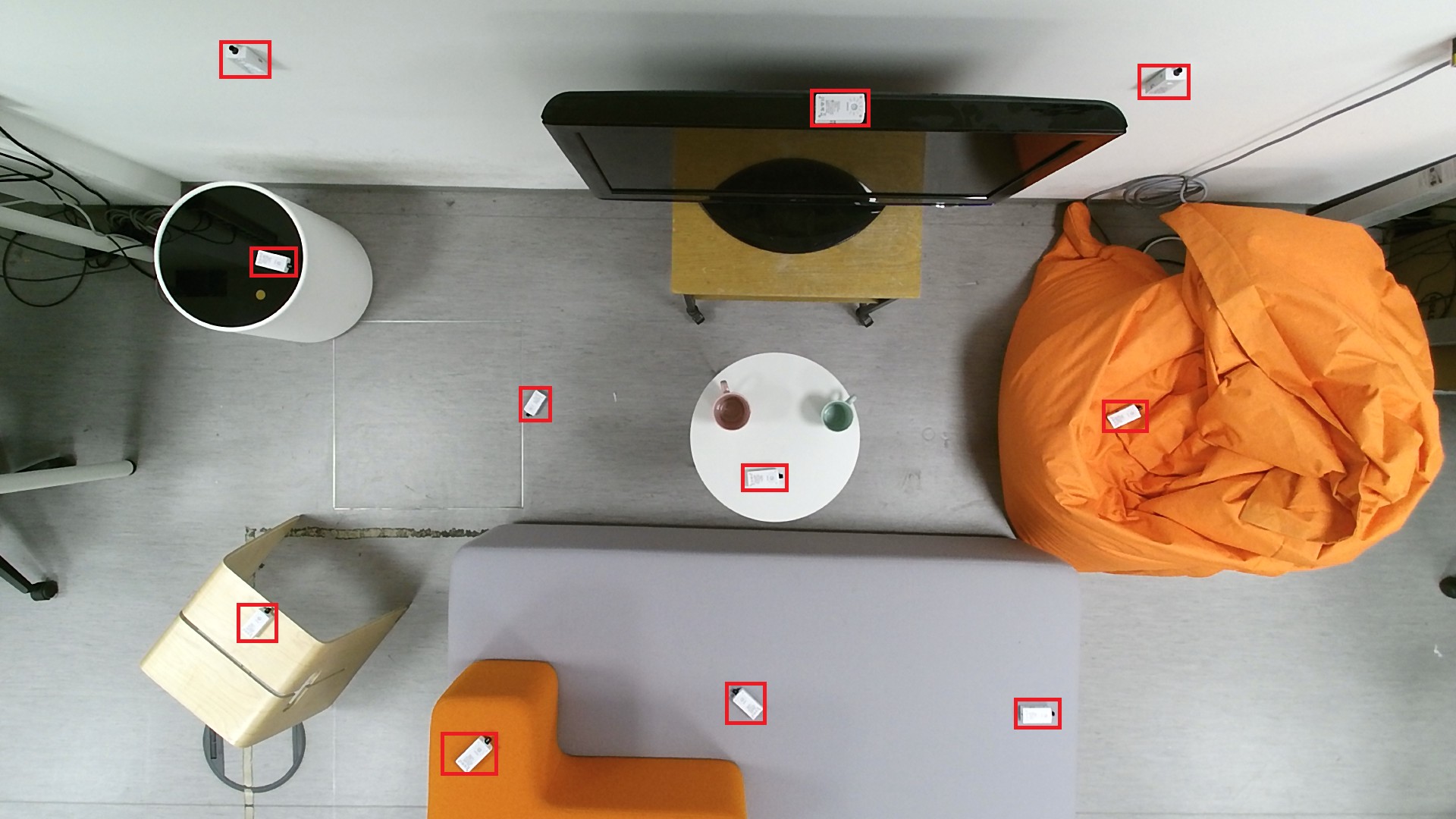}}{37pt}{70pt}\label{fig:room5}}\\
		\vspace{-20pt}
		\subfloat[][]{\topinset{{\footnotesize(g)}}{\includegraphics[width=0.49\linewidth]{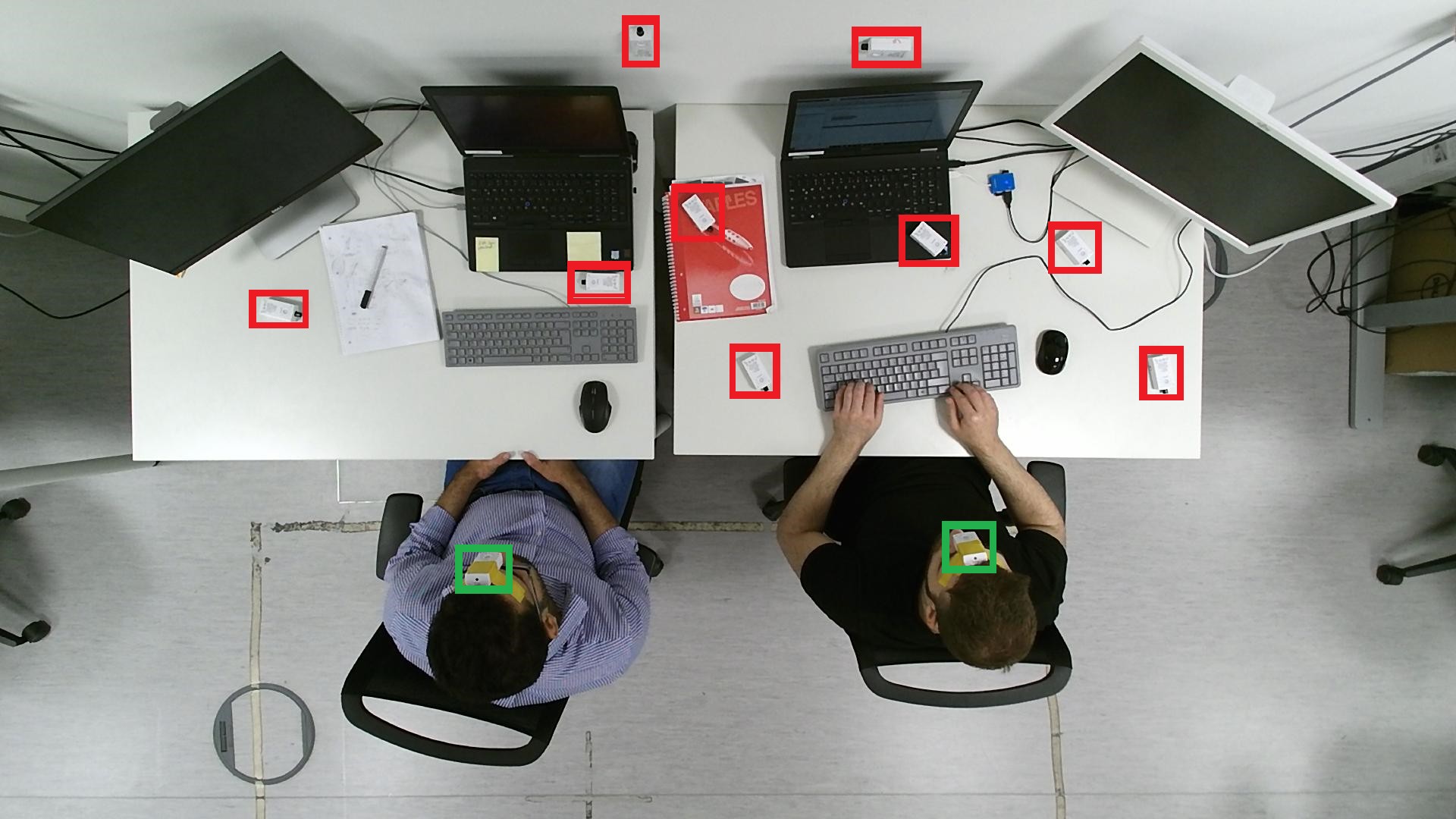}}{61pt}{113pt}\label{fig:room6}}
		\hspace{0.01em}
		\subfloat[][]{\topinset{{\footnotesize(h)}}{\includegraphics[width=0.49\linewidth]{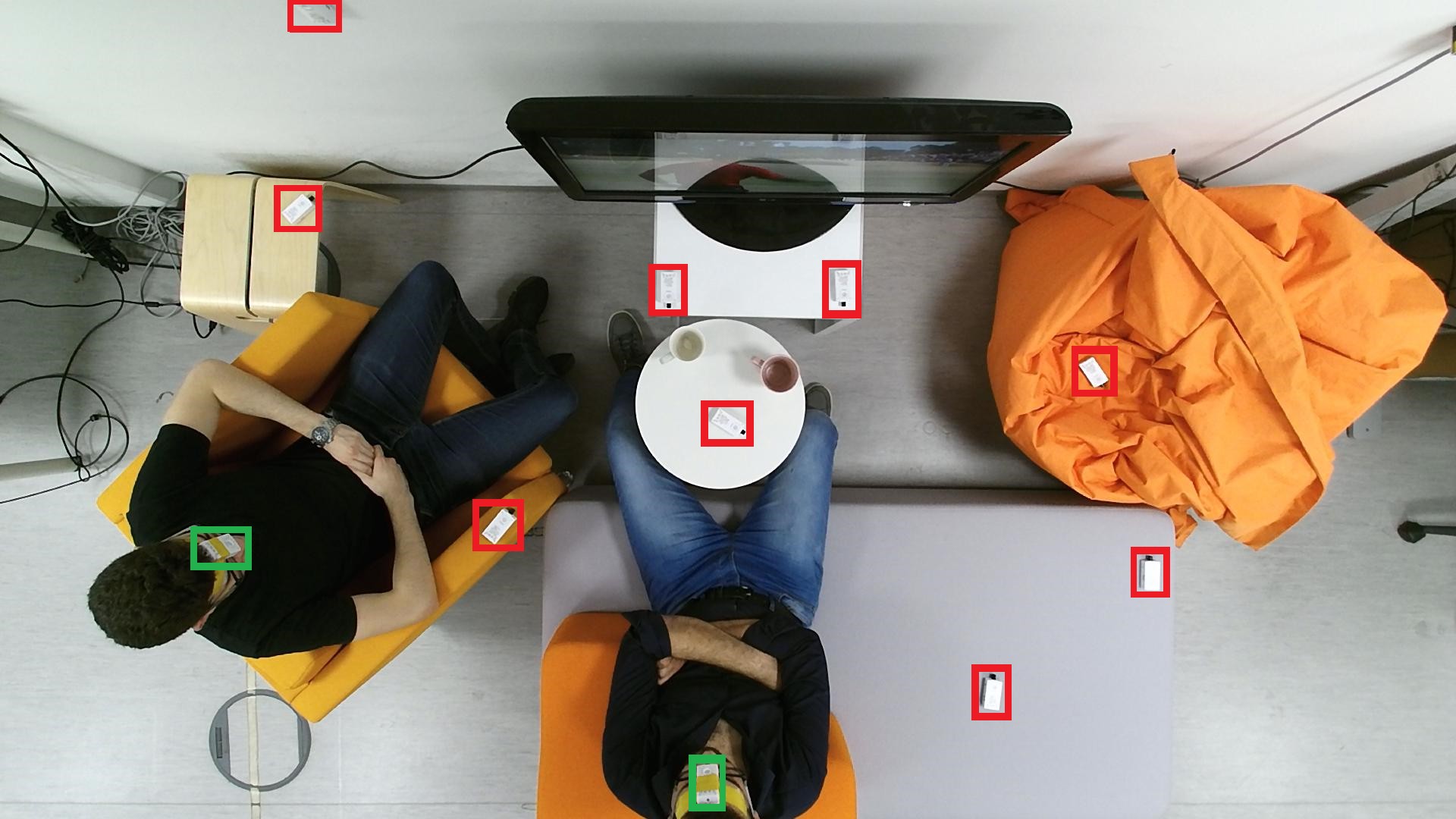}}{61pt}{113pt}\label{fig:room7}}
		% 		\hspace{0.01em}
		% 		\subfloat[][]
		% 		{\includegraphics[width=0.32\linewidth]{images/unit_sphere_over_cad.png}\label{fig:isocel_rays}}
% 		\vspace{5pt}
	\end{center}
	\vspace{-15pt}
	\caption[Room\_1 Full Scene]{
	Illustration of the 5 dataset scenarios (5 rooms). \textit{(a)} shows Room 1, from the top-view camera. \textit{(b)} and \textit{(c)} are detailed CAD models of Room 1, illustrating the luminaire positions (green boxes) and the scene subdivision into patches (\cf~\nameref{sec:scene_composition_analysis}) respectively. Room 2 (not shown) is similar to Room 1, but it has no panels across the central four desks. \textit{(d)}, \textit{(e)} and \textit{(f)} depict Rooms 3, 4, 5, also used to evaluate the illumination estimation quality (note the red boxes, highlighting the positions of the luxmeters for light measurement). \textit{(g)} and \textit{(h)} illustrate Rooms 4 and 5 additionally featuring human activity (note the green boxes on the occupants forehead, indicating the worn luxmeters).} \label{fig:rooms1_full_scene}
	\vspace{-5pt}
\end{figure}

\subsection{Dataset and Metrics}\label{sec:dataset}

% \usepackage{multirow}
\begin{table*}[!ht]
\centering
\caption[Illumination Estimation Error]{
Comparative evaluation of our approach \vs the Relux commercial software, alongside the other ablation studies referred to in Sec.~\ref{sec:luxcom}. 
%Relux is only tested in the case of manual input CAD model (``Relux w/ CAD''). We test it against ours, which does not need the CAD (\textit{last row}, ``Ours w/o CAD''), and against ours with the provided CAD (\textit{second row}, ``Ours w/ CAD'') for fairness. ``no\_LSC'' and ``no\_LDC'' stand for removing the light emitter and sensor distribution curves.  ``within camera FOV'' in the last two rows means cropping the 3D scene to the camera field-of-view (FOV). In the table, only ``Ours w/o CAD (within camera FOV)'' uses an estimated 3D scene geometry. Please refer to Sec.~\ref{sec:luxcom} for more details.
%``Avg. 1-8'' and ``Avg. 2-7'' are average errors in lux across the corresponding luxmeters. The latter refer to those luxmeter within sight of the depth camera (we can only estimate illumination at spots within the sensor field of view). Values for individual luxmeters, cols 1-8, correspond to the average lux values estimated for the different applied lighting combinations.
%
% \vspace{-10pt}
}
\label{table:quantitative_room1_2}
\resizebox{1\linewidth}{!}{\begin{tabu}{|c|c|c|c|c|c|c|c|c|c|c|l|c|c|c|c|c|c|c|c|c|c|}
\cline{1-11} \cline{13-22}
\multirow{3}{*}{\textbf{\begin{tabular}[c]{@{}c@{}}\\ \end{tabular}}}                                                                   & \multicolumn{10}{c|}{\textbf{\begin{tabular}[c]{@{}c@{}}Room 1\\ Avg. Spatial Light Estimation Error in \textit{lux}\end{tabular}}}                                                                                                                                          &  & \multicolumn{10}{c|}{\textbf{\begin{tabular}[c]{@{}c@{}}Room 2\\ Avg. Light Estimation Error in \textit{lux}\end{tabular}}}                                                                                                                                         \\ \cline{2-11} \cline{13-22} 
                                                                                             & \multicolumn{8}{c|}{\textbf{Luxmeters}}                                                               & \multicolumn{2}{c|}{\textbf{}}                                                                                                  &  & \multicolumn{8}{c|}{\textbf{Luxmeters}}                                                               & \multicolumn{2}{c|}{\textbf{}}                                                                                                 \\ \cline{2-11} \cline{13-22} 
                                                                                             & \textbf{1} & \textbf{2} & \textbf{3} & \textbf{4} & \textbf{5} & \textbf{6} & \textbf{7} & \textbf{8} & \textbf{\begin{tabular}[c]{@{}c@{}}Avg.\\ 1-8\end{tabular}}  & \textbf{\begin{tabular}[c]{@{}c@{}}Avg.\\ 2-7\end{tabular}}  &  & \textbf{1} & \textbf{2} & \textbf{3} & \textbf{4} & \textbf{5} & \textbf{6} & \textbf{7} & \textbf{8} & \textbf{\begin{tabular}[c]{@{}c@{}}Avg.\\ 1-8\end{tabular}} & \textbf{\begin{tabular}[c]{@{}c@{}}Avg.\\ 2-7\end{tabular}}  \\ \cline{1-11} \cline{13-22} 
\textbf{\begin{tabular}[c]{@{}c@{}}Relux w/ CAD\\ \ \end{tabular}}                                                                               & 167        & 96         & 27         & 26         & 43         & 10         & 96         & 39         & \begin{tabular}[c]{@{}c@{}}63\end{tabular}          & \begin{tabular}[c]{@{}c@{}}50\end{tabular}          &  & 206        & 97         & 27         & 80         & 97         & 49         & 73         & 44         & \begin{tabular}[c]{@{}c@{}}84\end{tabular}         & \begin{tabular}[c]{@{}c@{}}71\end{tabular}          \\ \cline{1-11} \cline{13-22} 
\textbf{\begin{tabular}[c]{@{}c@{}}Ours w/ CAD\\ \ \end{tabular}}                  & 69         & 24         & 22         & 38         & 28         & 28         & 38         & 41         & \textbf{\begin{tabular}[c]{@{}c@{}}36\end{tabular}} & \textbf{\begin{tabular}[c]{@{}c@{}}30\end{tabular}} &  & 70         & 57         & 76         & 106        & 75         & 69         & 55         & 53         & \textbf{\begin{tabular}[c]{@{}c@{}}70\end{tabular}}  & \begin{tabular}[c]{@{}c@{}}73\end{tabular}          \\ \cline{1-11} \cline{13-22} 
\textbf{\begin{tabular}[c]{@{}c@{}}Ours w/ CAD\\ (no\_LDC\_LSC)\end{tabular}}              & 188        & 150        & 33         & 45         & 43         & 34         & 91         & 65         & \begin{tabular}[c]{@{}c@{}}81\end{tabular}          & \begin{tabular}[c]{@{}c@{}}66\end{tabular}          &  & 207        & 114        & 99         & 148        & 105        & 117        & 93         & 81         & \begin{tabular}[c]{@{}c@{}}120\end{tabular}        & \begin{tabular}[c]{@{}c@{}}112\end{tabular}         \\ \cline{1-11} \cline{13-22} 
\textbf{\begin{tabular}[c]{@{}c@{}}Ours w/ CAD\\ (no\_LSC)\end{tabular}}                       & 199        & 152        & 29         & 41         & 40         & 33         & 95         & 57         & \begin{tabular}[c]{@{}c@{}}81\end{tabular}          & \begin{tabular}[c]{@{}c@{}}65\end{tabular}          &  & 213        & 117        & 82         & 125        & 97         & 97         & 86         & 63         & \begin{tabular}[c]{@{}c@{}}110\end{tabular}        & \begin{tabular}[c]{@{}c@{}}100\end{tabular}         \\ \cline{1-11} \cline{13-22} 
\textbf{\begin{tabular}[c]{@{}c@{}}Ours w/ CAD\\ (no\_LDC)\end{tabular}}                       & 73         & 45         & 24         & 32         & 40         & 34         & 46         & 52         & \begin{tabular}[c]{@{}c@{}}43\end{tabular}          & \begin{tabular}[c]{@{}c@{}}37\end{tabular}          &  & 69         & 80         & 98         & 136        & 70         & 84         & 56         & 62         & \begin{tabular}[c]{@{}c@{}}82\end{tabular}         & \begin{tabular}[c]{@{}c@{}}87\end{tabular}          \\ \cline{1-11} \cline{13-22}  \cline{1-11} \cline{13-22} 
\textbf{\begin{tabular}[c]{@{}c@{}}Ours w/ CAD\\ (within camera FOV) \end{tabular}} & -          & 64         & 28         & 20         & 17         & 22         & 52         & -          & -                                                              & \begin{tabular}[c]{@{}c@{}}34\end{tabular}          &  & -          & 54         & 36         & 59         & 101        & 69         & 54         & -          & -                                                             & \textbf{\begin{tabular}[c]{@{}c@{}}62\end{tabular}} \\ [-2.5pt] \tabucline[1.5pt]{1-11} \tabucline[1.5pt]{13-22}
\textbf{\begin{tabular}[c]{@{}c@{}}Ours w/o CAD\\(within camera FOV)\end{tabular}}                   & -          & 53         & 41         & 67         & 68         & 40         & 98         & -          & -                                                              & \begin{tabular}[c]{@{}c@{}}61\end{tabular}          &  & -          & 98         & 90         & 85         & 136        & 108        & 77         & -          & -                                                             & \begin{tabular}[c]{@{}c@{}}99\end{tabular}          \\ \cline{1-11} \cline{13-22} 
\end{tabu}}
% \vspace{-8pt}
\end{table*}

% There is currently no ground-truthed dataset for benchmarking light measurements and relating it to human perception. 
In addition to the data of \cite{tsesmelis2018b,tsesmelis2018a} we provide a dataset with RGBD images and light measurements for 3 new and diverse scenes with various new illumination scenarios (combinations of on/off light sources for each scene). The new dataset includes offices, meeting rooms and resting areas as depicted in Fig. \ref{fig:rooms1_full_scene}. For two of the scenes (\cf Figs. \ref{fig:rooms1_full_scene}b,c) detailed 3D CAD models are available, as well as an accurate labelling of the object textures and reflectivities. This allows the comparison to ReLux, a state-of-the-art commercial software for light modelling. Moreover, for two other scenes, we provide annotated human activity (watching TV, working at the desk, chatting, \etc), detailing the people position and their visual frustum of attention (VFOA) -- \cf Figs. \ref{fig:rooms1_full_scene}g,h -- as well as the illumination reaching their sight (\ie within in their VFOA). This allows benchmarking lighting efficiency, in relation to human perception.

For each frame, the illumination maps are ground-truthed by a number (8-11) of synchronized luxmeters (depicted in Fig.~\ref{fig:rooms1_full_scene} as red boxes), to measure the light intensity reaching the specific spot, in \emph{lux}. %Depending on the scene, we installed between 8 and 11 luxmeters. 
We set to measure the light estimation quality by the absolute error in \textit{lux}, compared to the corresponding luxmeter readings.

To assess the quality of the human-centric system, we benchmark people detection and (quantized) pose estimation accuracy by the established metrics of mean average precision (mAP) and classification accuracy. Both measures are quantified on the external larger dataset of \cite{DemirkusetalVISAPP17TopViewDet} where people and their head pose are labelled. %This includes 123900 images (89180 train + 34720 test), mainly featuring a single person from a top-view camera. All frames are labelled with bounding boxes of people. Here we additionally label all frames with the people head pose. 
On the other hand, we assess the accuracy in estimating the perceived illumination by each person with additional luxmeters, worn by the users on their forehead (see green boxes in Figs.~\ref{fig:rooms1_full_scene}g,h). This is also quantified by the absolute errors in \textit{lux}.

%Depending on the scene we use $8$ or $9$ luxemetes for evaluating the spatial illumination across the environment and $2$ luxmeters for measuring the light intenisty that arrives to each one of the occupants appearing in the two extended scenes. For the evaluation of the ambient illumination we chose mainly locations over desk areas, since they are of major importance in the lighting field. For each luxmeter, we additionally report the type and their specific light sensitivity characteristic curve (LSC, see Fig. \ref{fig:ldc_lsc}), namely the sensor sensitivity across the incident light angles.
%
% Moreover, for each room we evaluated multiple different light activation scenarios (luminaires switched on or off), see Fig. \ref{fig:illumination_combinations_room1} for a sample of  images obtained from room\_1. We target the use of rgb and depth input just for light measurement, the use of luxmeters as ground truth, and all other provided information for ablation studies.
%
% \begin{figure}[ht!]
% 	\begin{center}
% 		\includegraphics[width=1\linewidth]{pics/illumination_combinations1a.png}
% 	\end{center}
% 	\caption[Light Activations - Room\_1]{Illustration of 4 illumination variants within room\_1. From the left to the right, the images illustrate the illumination provided by 1, 3, 4 and all 8 luminaires switched on in the scene.}\label{fig:illumination_combinations_room1}
% % 	\vspace{-15pt}
% \end{figure}

\subsection{Ambient Light Estimation}\label{sec:luxcom}

Table~\ref{table:quantitative_room1_2}, first compares our proposed light estimation approach (``Ours w/o CAD'') to the commercial software Relux, within the scenes of Rooms 1 and 2. %Since Relux requires the manual input CAD design, we also test our model when provided it (``Ours w/ CAD''), as well as a number of ablation studies.
Given the CAD model, our method (``Ours w/ CAD'') outperforms the Relux software (``Relux w/ CAD'') on both rooms. We achieve an average error of $36$ and $70$ \textit{lux} for Rooms 1 and 2 across the installed luxmeter sensors (``Avg.\ 1-8''), \vs the Relux errors of $63$ and $84$ \textit{lux} respectively. %Furthermore, the computational complexity of our method allows real-time performance, while each Relux simulation requires over 20 minutes. In general, one may note that Room 2 is more difficult than Room 1, because the removed panels among the central desks (\cf caption of Fig.~\ref{fig:rooms1_full_scene}) result in light rays reaching each scene part from nearly each luminaire.
Note also that these errors are to be intended over an illumination range of [0,2000] \textit{lux}. In both cases, these are good estimates for commissioning, since errors below 200 \textit{lux} are generally acceptable in the lighting industry.
%Lighting can reach up to the range of 2K for Room 1, 2, 5 (\textit{dynamic}), 1K for Room 4, 5, 4 (\textit{dynamic}) and 500 lux for Room 3 setting these errors in a negligible range. 

Table~\ref{table:quantitative_room1_2} also contains ablation studies ``w/ CAD'', given by removing the LSC and LDC distribution curves (``no\_LDC\_LSC'', ``no\_LDC'', ``no\_LSC''). Results confirm that both LDC and LSC are key features to best performance. It is of interest that LSC yields a larger error reduction than the LDC for both rooms. In our view, this happens because LSC properly considers the angle of impacting rays, especially down-weighting the rays coming to the sensor from the side, reflecting the sensor absorption characteristic. %This benefits less in the case of LDC, as those rays emitted laterally by the luminaires are only reaching the high portion of walls (\eg above $1.7m$), where we did not measure lighting.

%The last row of Table~\ref{table:quantitative_room1_2} reports performance of our full light estimation approach, which does not require any manual input CAD (``Ours w/o CAD''). 
When only using the RGBD image (\ie ``Ours w/o CAD''), we may only estimate light propagation to those scene parts which are within the camera field-of-view (FOV). Since the camera is placed around the room center, we leave luxmeters 1 and 8 at the scene corners out (\cf Fig.~\ref{fig:rooms1_full_scene}a), and only provide the average errors for luxmeters 2-7 (61 and 99 \textit{lux} for room 1 and 2 respectively). 
% Dealing with the estimated room geometry instead of the manual CAD 
This case is clearly more challenging, partly because of the camera FOV, and partly because the room reconstruction is effectively a noisy 2.5D. The camera FOV only covers $\sim$40\% of the actual scene, as it misses the ceiling and most of the walls. Moreover, the camera-based scene reconstruction only allows a noisy estimation of those surfaces which are visible from the camera viewpoint. We set to discern which factor matters the most by the ablation method ``Ours w/ CAD (within camera FOV)'', which considers a manual input CAD, cropped according to the camera FOV. As shown in Table~\ref{table:quantitative_room1_2}, this performs closely to ``Ours w/ CAD'' (the slightly lower error for room 2 seems due to the challenging scene with entangled lights, where removing walls reduces some of the possibly misestimated light rays). We conclude that most of the added challenge comes from the noise in the point-cloud and in the effective 2.5D estimate, resulting in a geometry with holes. However, in contrast to all other techniques, the proposed ``Ours w/o CAD'' is the sole fully-automatic.

The left section of Table~\ref{table:spatial_human_light_percetion} illustrates similar performance of our full automatic technique on the rooms 3, 4 and 5, depicted in Fig.~\ref{fig:rooms1_full_scene}. Additionally, we introduce two sets of \emph{dynamic} experiments in rooms 4 and 5. By \emph{dynamic} we mean the inclusion of people activities with different VFOAs and interactions within the scenes. As it can be seen results are in line with previous experiments.

\begin{table*}[!ht]
\centering
\caption[Average Estimated Illumination Error]{Values on the left present the spatial light estimation errors across rooms 3-5, obtained with the proposed approach (w/o CAD). \emph{Dynamic} scenes are additional image sets, including people, activities and scene interactions. (See Sec.~\ref{sec:luxcom} for details.) Values on the right table side are the error in estimating the light perception of people with our proposed method, compared to the ground-truthed values provided by head-worn luxmeters. We also report error values for the cases where the head pose is provided (\emph{oracle}). %Overall the error accounted for by the head pose estimator is significant, compared to the quality of the light estimation.
% \vspace{-10pt}
}
\label{table:spatial_human_light_percetion}
\resizebox{1\linewidth}{!}{\begin{tabular}{|c|c|c|c|c|c|c|c|c|c|c|c|c|c!{\vrule width2pt}c|c|c|c|c|}
\hline
\multicolumn{2}{|c|}{\multirow{3}{*}{\textbf{}}}                                                                                                                         & \multicolumn{12}{c!{\vrule width2pt}}{\textbf{Avg.  Spatial  Light  Estimation  Error  in \textit{lux}}}                                                                                                                                                                                      & \multicolumn{4}{c|}{\textbf{Avg.  Human  Light  Perception  Error  in \textit{lux}}}                                                                   \\ \cline{3-18} 
\multicolumn{2}{|c|}{}                                                                                                                                                   & \multicolumn{11}{c|}{\textbf{Luxmeters}}                                                                                                                                                                                                      & \textbf{}           & \multirow{2}{*}{\textbf{}}   & \multicolumn{2}{c|}{\textbf{\begin{tabular}[c]{@{}c@{}}Luxmeters\\ (head-worn)\end{tabular}}} & \textbf{}      \\ \cline{3-14} \cline{16-18} 
\multicolumn{2}{|c|}{}                                                                                                                                                   & \textbf{1}          & \textbf{2}          & \textbf{3}          & \textbf{4}          & \textbf{5}          & \textbf{6}          & \textbf{7}          & \textbf{8}          & \textbf{9}          & \textbf{10}        & \textbf{11}        & \textbf{Avg.}       &                              & \textbf{1}                                    & \textbf{2}                                    & \textbf{Avg.}  \\ \hline
\multirow{3}{*}{\rotatebox[origin=c]{90}{\textbf{\begin{tabular}[c]{@{}c@{}}Static\\ scene\end{tabular}}}}  & \textbf{Room 3}                                                                      & 70                  & 93                  & 69                  & 23                  & 25                  & 28                  & 59                  & 28                  & 49                  & 71                 & 82                 & 54                  & -                            & -                                             & -                                             & -              \\ \cline{2-18} 
                                                                                  & \textbf{Room 4}                                                                      & 18                  & 23                  & 41                  & 26                  & 76                  & 23                  & 35                  & 34                  & 40                  & 69                 & 31                 & 38                  & -                            & -                                             & -                                             & -              \\ \cline{2-18} 
                                                                                  & \textbf{Room 5}                                                                      & 35                  & 38                  & 23                  & 27                  & 29                  & 49                  & 23                  & 40                  & 33                  & 25                 & 34                 & 32                  & -                            & -                                             & -                                             & -              \\ \hline
\multirow{4}{*}{\rotatebox[origin=c]{90}{\textbf{\begin{tabular}[c]{@{}c@{}}Dynamic\\ scene\end{tabular}}}} & \multirow{2}{*}{\textbf{\begin{tabular}[c]{@{}c@{}}Room 4\\ (\textit{dynamic})\end{tabular}}} & \multirow{2}{*}{62} & \multirow{2}{*}{26} & \multirow{2}{*}{68} & \multirow{2}{*}{65} & \multirow{2}{*}{48} & \multirow{2}{*}{57} & \multirow{2}{*}{44} & \multirow{2}{*}{30} & \multirow{2}{*}{28} & \multirow{2}{*}{-} & \multirow{2}{*}{-} & \multirow{2}{*}{48} & \textbf{Est. head pose}      & 216                                           & 166                                           & 191            \\ \cline{15-18} 
                                                                                  &                                                                                      &                     &                     &                     &                     &                     &                     &                     &                     &                     &                    &                    &                     & \textbf{Oracle head pose}    & 98                                            & 92                                            & 95             \\ \cline{2-18} 
                                                                                  & \multirow{2}{*}{\textbf{\begin{tabular}[c]{@{}c@{}}Room 5\\ (\textit{dynamic})\end{tabular}}} & \multirow{2}{*}{35} & \multirow{2}{*}{34} & \multirow{2}{*}{44} & \multirow{2}{*}{20} & \multirow{2}{*}{32} & \multirow{2}{*}{40} & \multirow{2}{*}{24} & \multirow{2}{*}{28} & \multirow{2}{*}{27} & \multirow{2}{*}{-} & \multirow{2}{*}{-} & \multirow{2}{*}{31} & \textbf{Est. head pose}      & 55                                            & 152                                           & 104            \\ \cline{15-18} 
                                                                                  &                                                                                      &                     &                     &                     &                     &                     &                     &                     &                     &                     &                    &                    &                     & \textbf{Oracle head pose}    & 42                                            & 69                                            & 55             \\ \hline
\end{tabular}}
\end{table*}

\subsection{People Detection and Head-pose Estimation}\label{sec:head_estimation_eval}

We attain best person detection performance by pre-training the detector on the large MS-COCO dataset%~\cite{lin2014microsoft} 
(80k training + 35k validation images) and then fine-tuning on a selection of diverse top-view images from \cite{DemirkusetalVISAPP17TopViewDet}. Following~\cite{DemirkusetalVISAPP17TopViewDet}, we select diverse frames by sampling one every 20 frames, which yields 4459 training + 1736 test images. On this test set, we achieve 98\% mAP (IoU=0.5).
%randomly partitioned the data into training and testing set, in a ratio of 70\%-30\% respectively.
% \todo{Irtiza: Please address the following points: i.\ verify the exact number, as obtained in the paper scenario; ii.\ check that you actually test detection on that dataset and not on the rooms 5 and 6, else change text here and in IVa.}

For estimating the VFOA, we train the tiny head-pose estimator of \cite{hasan2017tiny} on the above-named selection of training images from \cite{DemirkusetalVISAPP17TopViewDet}, after having labelled all dataset.
We test on head-pose angles quantized into 4 and 8 classes and obtained accuracy of 70.7\% and 43.2\% respectively on our test set. 
%Out of visual inspection of the corresponding confusion matrices in Fig.~\ref{fig:confMat}, it shows that 
The degraded performance in the 8-class case is justified due to adjacent viewing angle confusion. This can be further explained by the \emph{tiny} head region of people in the images, of just $40\times50$ pixels. Thus, we considered the 4-quantized-angle head pose estimator.

\subsection{Human-centric Light Estimation} \label{sec:light_perception}

On the right part of Table~\ref{table:spatial_human_light_percetion}, we illustrate the error of our method in estimating the amount of illumination arriving at the people sight. We assume as ground truth the illumination estimation of a luxmeter which the occupants wear on their forehead. Average errors for rooms 4 and 5 are 191 and 104 \textit{lux} respectively. The figures are relatively large, compared to the average light estimation errors reported for the spatial light estimation. We justify this due to challenges in the light propagation as well as due to the error in the head pose estimation. We set therefore to estimate errors when the head pose is given by an oracle.

Light estimation errors for the case of oracle head-pose are also reported in the same section of Table~\ref{table:spatial_human_light_percetion} and are significantly lower ($\sim50\%$). This shows that there is much progress needed in head pose estimation. Still, the residual error remains larger in average, compared to the scene illumination estimates (95 \vs 48 \textit{lux} for Room 4; 55 \vs 31 \textit{lux} for Room 5). We explain the discrepancy by two main factors: \textbf{a}) the reconstructed 3D mesh is less accurate for the people heads than on desks; and \textbf{b}) the light estimation arriving at the people suffers from the limited FOV of the depth sensor, since it excludes parts of walls, an important factor for the head-worn luxmeter facing them.

\subsection{Human-centric Light Management System} \label{sec:applications}

\begin{figure}[!ht]
    % \vspace{-5pt}
	\begin{center}
	    \includegraphics[width=1\linewidth]{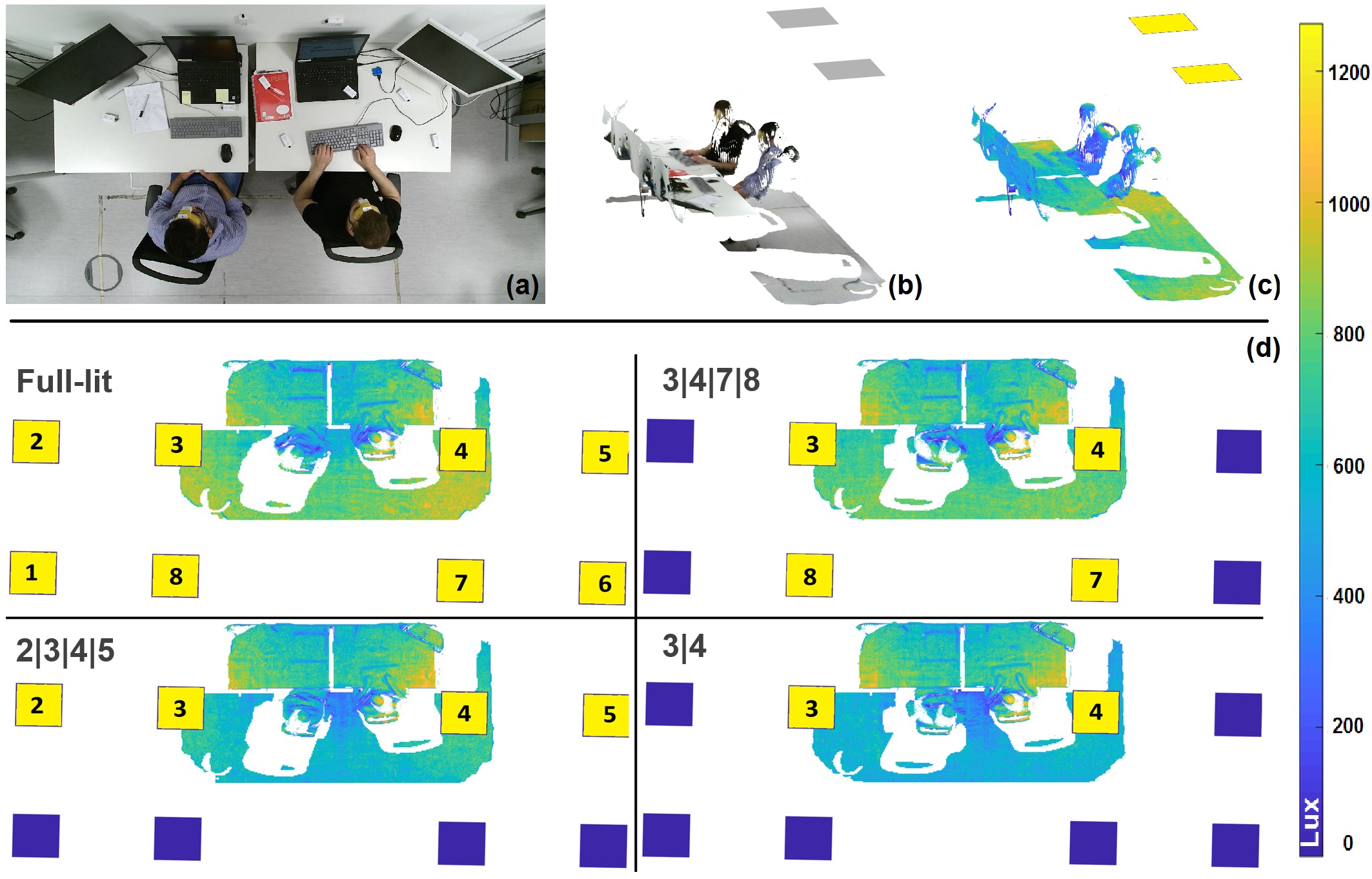}
	\end{center}
	\caption[ILS Qualitative Analysis]{
	Illustration of the scene, person activity and illumination map for room 4.
	\textit{(a)} shows the top-view image, corresponding to the 3D in \textit{(b)} with mapped textures.
	\textit{(c)} and \textit{(d)} present illumination maps, from 3D- and top-view respectively.
	In more detail, \textit{(d)} shows four lighting setups, from the full lit (1$\mid$2$\mid$3$\mid$4$\mid$5$\mid$6$\mid$7$\mid$8) to the most energy-saving scenario (3$\mid$4). %Note how the illumination map over the space in front of the people (desks, monitors) is only minimally affected across the setups, which justifies proposing the invisible light switch.
	} \label{fig:test_power_saving}
% 	\vspace{-10pt}
\end{figure}

Table~\ref{table:power_saving} evaluates whether we can effectively adopt our proposed end-to-end system, \ie \textit{ILS}, for light management, implementing the \emph{``invisible light switch''} principle. The table reports experiments for rooms 4 and 5, whereby occupants engage in activities, while we change the room illumination and switch off some of the 8 luminaires (setup 1$\mid$2$\mid$3$\mid$4$\mid$5$\mid$6$\mid$7$\mid$8 refers to ``full-lit'' where all luminaires are on, while 3$\mid$4 means keeping only luminaires 3 and 4 on. Fig.~\ref{fig:test_power_saving} illustrates the setup of room 4 and the resulting illumination maps, under different lighting setups.

The main performance measure in Table~\ref{table:power_saving} is $\varDelta_{watt}$, which quantifies how much power can be saved when turning some luminaires off. For example, when keeping only luminaires 3 and 4 on (setup 3$\mid$4), one may save up to 580.8 \textit{watt}. Over a full working day (8 hours), this setup allows to save up to 
% 515.8 watt/hour (including the power consumption of the processing unit) which is approx. 
99 \textit{KWh} (including the power consumption of the processing unit)%\todo{Theo: how did you get this figure? Also, you had left ``hour'' out. third point: why reporting this so accurately? It seems beyond the system precision. Should it be 12.3 K ? One more point: please also double check the ``K'' or whether it should be watt hour just}
, meaning 66\% of energy efficiency.

% \usepackage{multirow}
\begin{table}[!ht]
\centering
\caption[ILS Quantitative Analysis]{Quantitative evaluation of how much power may be saved by \textit{ILS} by switching off luminaires that are not directly affecting the human light perception of the occupants, in rooms 4 and 5. %The table additionally reports the difference in the perceived illumination in \textit{lux} $\varDelta_{lux}$, measured by luxmeters worn on the occupants' forehead. Finally, the table reports the error in \textit{lux} when estimating the illumination in sight of each person, which results in an end-to-end solution.
%as for running the system as an end-to-end solution. 
%See also Fig.~\ref{fig:test_power_saving} for illustration.
% \vspace{-5pt}
}
\label{table:power_saving}
\resizebox{1\linewidth}{!}{\begin{tabular}{|c|c|c|c|c|c|c|c|c|c|}
\hline
\multicolumn{3}{|c|}{}                                                                                                                              & \multicolumn{3}{c|}{\textbf{Room  4  (\textit{dynamic})}}   & \multicolumn{4}{c|}{\textbf{Room  5  (\textit{dynamic})}}                          \\ \hline
\multicolumn{3}{|c|}{\textbf{\begin{tabular}[c]{@{}c@{}}Active  luminaires\\ (out  of  8  available,  1$\mid$2$\mid$3$\mid$4$\mid$5$\mid$6$\mid$7$\mid$8)\end{tabular}}}               & \textbf{3$\mid$4$\mid$7$\mid$8} & \textbf{2$\mid$3$\mid$4$\mid$5} & \textbf{3$\mid$4} & \textbf{1$\mid$2$\mid$3$\mid$4$\mid$5$\mid$6} & \textbf{2$\mid$3$\mid$4$\mid$5} & \textbf{1$\mid$3$\mid$4$\mid$6} & \textbf{3$\mid$4} \\ \hline
\multicolumn{3}{|c|}{\textbf{\begin{tabular}[c]{@{}c@{}}$\varDelta_{watt}$\\ (\wrt full-lit)\end{tabular}}}                                                                                                          & 387.2            & 387.2            & 580.8        & 193.6                & 387.2            & 387.2            & 580.8        \\ \hline
\multirow{6}{*}{\rotatebox[origin=c]{90}{\textbf{\begin{tabular}[c]{@{}c@{}}Luxmeters \\ (head-worn)\end{tabular}}}} & \multirow{2}{*}{\textbf{1}} & \textbf{\begin{tabular}[c]{@{}c@{}}$\varDelta_{lux}$\\ (\wrt full-lit)\end{tabular}}       & 116.15           & 123.77           & 189.01       & 106.52               & 148.12           & 157.07           & 191.15       \\ \cline{3-10} 
                                                                                          &                             & \textbf{\begin{tabular}[c]{@{}c@{}}Light est. error\\ (\wrt GT)\end{tabular}} & 167.2           & 144.09           & 102.73       & 22.94                & 12.97            & 13.59            & 25.69        \\ \cline{2-10} 
                                                                                          & \multirow{2}{*}{\textbf{2}} & \textbf{\begin{tabular}[c]{@{}c@{}}$\varDelta_{lux}$\\ (\wrt full-lit)\end{tabular}}       & 97.68            & 125.15           & 169.72       & 99.17                & 154.28           & 167.93           & 194.85       \\ \cline{3-10} 
                                                                                          &                             & \textbf{\begin{tabular}[c]{@{}c@{}}Light est. error\\ (\wrt GT)\end{tabular}} & 194.63           & 171.74           & 131.55       & 9.4                  & 241.12           & 2.81             & 203.69       \\ \hline
\end{tabular}}
\end{table}

The second important measure in the table is $\varDelta_{lux}$, which quantifies how much the illumination in sight changes. We measure this by means of luxmeters which the occupants wear on their forehead, as also previously described in Sec.~\ref{sec:light_perception}. For example, in room 4 and setup 3$\mid$4, the person on the right (\cf Fig.~\ref{fig:test_power_saving}) wearing the head-worn luxmeter 1 perceives a difference of 189.01 \textit{lux} (over the ambient lighting of 1200 \textit{lux}), the largest in the most-energy saving scenario.
%Still, this value can be considered a minor variation in regards to the overall ambient lighting in the scene which for this specific scenario is 1200 lux.
%, an agreed threshold for noticeable differences in lighting installations.

Finally, we report the light estimation error. In the above example (room 4, occupant 1, setup 3$\mid$4) the error is 102.73 \textit{lux}. This remains comparable to the actual light variation $\varDelta_{lux}$. Fig.~\ref{fig:test_power_saving} supports the results qualitatively. 
%In more detail, in Fig.~\ref{fig:test_power_saving}d we illustrate the estimated illumination maps across the four lighting scenarios, from the full lit (1$\mid$2$\mid$3$\mid$4$\mid$5$\mid$6$\mid$7$\mid$8) to the most energy-saving (3$\mid$4). The little change over the space in front of the people (desks, monitors) substantiates adopting the proposed invisible light switch in light management systems.

\section{Conclusion} \label{sec:conclusion}

We have motivated, introduced and benchmarked the first real-time, end-to-end and human-centric light management system. \textit{ILS} is based on models for estimating the scene lighting as well as the illumination in sight of the scene occupants. Based on these, we have defined an invisible light switch light management algorithm, to switch off or dim luminaires which are partially visible. Key aspects of this proposition are the relatively small light estimation errors, compared to lighting industry standards, which support the proposed \textit{``invisible light switch''}.
\begin{acks}
This project has received funding from the European Union's Horizon 2020 research and innovation programme under the Marie Sklodowska-Curie Grant Agreement No. 676455 and has been partially supported by the project of the Italian Ministry of Education, Universities and Research (MIUR) "Dipartimenti di Eccellenza 2018-2022".
\end{acks}

\bibliographystyle{SageH}
\bibliography{bibliography}

\end{document}